# SATVSR: Scenario Adaptive Transformer for Cross Scenarios Video Super-Resolution

Yongjie Chen[1,a], Tieru Wu[1*]

[1]School of Artificial Intelligence, Jilin University, Changchun, China

[a]Email: yjchen20@mails.jlu.edu.cn, [*]Email: wutr@jlu.edu.cn

**Abstract.** Video Super-Resolution (VSR) aims to recover sequences of high-resolution (HR) frames from low-resolution (LR) frames. Previous methods mainly utilize temporally adjacent frames to assist the reconstruction of target frames. However, in the real world, there is a lot of irrelevant information in adjacent frames of videos with fast scene switching, these VSR methods cannot adaptively distinguish and select useful information. In contrast, with a transformer structure suitable for temporal tasks, we devise a novel adaptive scenario video super-resolution method. Specifically, we use optical flow to label the patches in each video frame, only calculate the attention of patches with the same label. Then select the most relevant label among them to supplement the spatial-temporal information of the target frame. This design can directly make the supplementary information come from the same scene as much as possible. We further propose a cross-scale feature aggregation module to better handle the scale variation problem. Compared with other video super-resolution methods, our method not only achieves significant performance gains on single-scene videos but also has better robustness on cross-scene datasets.

## 1 Introduction

Video super-resolution (VSR) is the task of enhancing and reconstructing low-resolution (LR) video frames into high-resolution video (HR) frames. The super-resolution results not only get the resolution that matches the target but also visually preserve more detailed textures in the image to achieve better results. At present, VSR has been deployed in some computer vision applications, such as video surveillance, satellite imagery, medical image reconstruction, ultra-high-definition display devices, etc. VSR technology can improve recognition accuracy, reduce hardware costs, and improve the viewing experience, which has attracted extensive attention from academia and industry in recent years.

Most notably, VSR requires aggregating information from a series of highly correlated but unaligned video frames. Most of the VSR methods [1] employ the following pipelines: propagation, alignment, aggregation, and upsampling. In BasicVSR [1], the advantages and disadvantages of various options under each component are studied, facilitating reproducible and fair comparisons. As a sequence modeling problem, using the Transformer framework to model the long-term dependencies of the input sequence is a suitable solution. In the VSR Transformer [2], the author uses optical flow for motion estimation and warps reference frame to align to target frame, and then designs a spatial-temporal attention layer, which achieves remarkable results when the number of frames is limited.

Chan et al. [1] show that long-term information is beneficial for video restoration. However, for real-world VSR applications, the LR videos are with unknown degradation and fast scene transitions.

In the previous work [3], more attention was paid to the processing of unknown degraded videos, which often achieved remarkable success in a limited environment with a single scenario and slow-motion video. For videos with rapid scenario switching, there are a large number of weakly correlated video frames, which may lead to exaggerated artifacts due to the error accumulation of irrelevant information during the propagation process. The designs upon overly stable scenarios cannot be generalized to complex scenarios switching environments in the wild. Therefore, this paper analyses the problem of cross-scenario in the real world. We prove that existing methods have a certain performance degradation for cross-scenario VSR, and we propose a spatial-temporal adaptive attention Transformer to achieve effective video representation learning for VSR.

In this study, we propose a **S**cenarios **A**daptive **T**ransformer for VSR (SATVSR) tasks, which effectively enhances the video reconstruction quality of scene-transformed. Different from traditional global attention applied to all video frame patches, SATVSR uses pre-trained optical flow models to mark the position of the patch in the target frame on other support frames and only calculate the similarity of the marked patches. We select the most similar patch as the information supplement of the target patch so that each patch in the target frame receives the information supplement from the same scene as possible. Furthermore, to better deal with the scale change problem, we design a cross-scale non-local aggregation module. We proposed SATVSR not only introduces a new attention mechanism, but also provides an efficient solution on cross-scenario problems.

The main contributions of this paper are summarized as follows:
- Instead of using the feature map after optical flow alignment, we use them as markers to filter out the most relevant features for attention map calculation. This design provides a novel opinion of reducing the complexity of the Transformer structure.
- We propose a Scenario Adaptive Transformer for the super-resolution of scene-transformed videos. Its efficient extraction of spatial-temporal information in similar scenes makes our method more robust to cross-scenario videos.
- Extensive experiments show that using the attention mechanism and cross-scale feature aggregation module, SATVSR can significantly outperform existing SOTA methods on several benchmark datasets.

## 2 Related Work

*2.1 Video Super-Resolution*
Several studies [3][5] indicate that adequate and proper utilization of inter-frame information has greatly influenced performance. The existing VSR methods can be divided into aligned methods and non-aligned methods. Among the methods with explicit alignment, motion estimation and motion compensation (MEMC) technology based on the optical flow are mostly used. The method [6][7] first uses classical algorithms to obtain optical flow, designs feature extraction and aggregation modules, and builds a network for high-resolution image reconstruction. However, given the occlusion and large motion scenarios, it is difficult to obtain accurate optical flow. DUF [5] circumvents this problem by using dynamic upsampling filters combined with 3D convolution to extract spatial-temporal information through implicit motion compensation. Although this method is simple, its computational cost tends to be high.

In the dynamic convolution-based research branch, EDVR [3] utilizes dynamic convolution for implicit alignment and designs pyramid and cascade architectures to handle large motions. However, such sliding-window-based methods cannot utilize textures on relatively distant frames. Recurrent structure-based methods [1] use the hidden state in the previous frame to transmit relevant information to achieve long-term modeling ability. Different from the single setting of the above methods, we dig the dependencies in the sequence and introduce a model for adaptively processing scene information.

*2.2 Vision Transformer*
In recent years, benefit to the great success of the Transformer structure and its variants in the field of natural language processing (NLP), many scholars have migrated the Transformer to computer vision tasks. The seminal work Vision Transformer (ViT) computes the attention between image patches, and has achieved impressive results in high-level vision image classification tasks. The variant T2T mines important local structures such as edges between adjacent pixels by recursively aggregating adjacent tokens into one token and gradually structuring images into tokens. IPT proposes a pre-trained model under low-level vision, which is effectively used for the desired task after fine-tuning, demonstrating long-term dependence on strong intrinsic feature capture ability.

Since the video sequence has a large number of patches, the computational cost under the standard Transformer architecture is increased. In order to reduce Transformer complexity in video sequence tasks, TimeSformer [9] explored several self-attention designs with different complexity. In the VSR task, VSR-Transformer [2] tries to use the attention mechanism to align different frames, which achieves ideal results when the number of frames is limited. Furthermore, TTVSR [10] improves the long-term modeling capability of VSR using a trajectory-aware Transformer. However, few related works combine attention mechanisms with dataset properties, and in this paper, we explore a method that is more robust to dataset transformations.

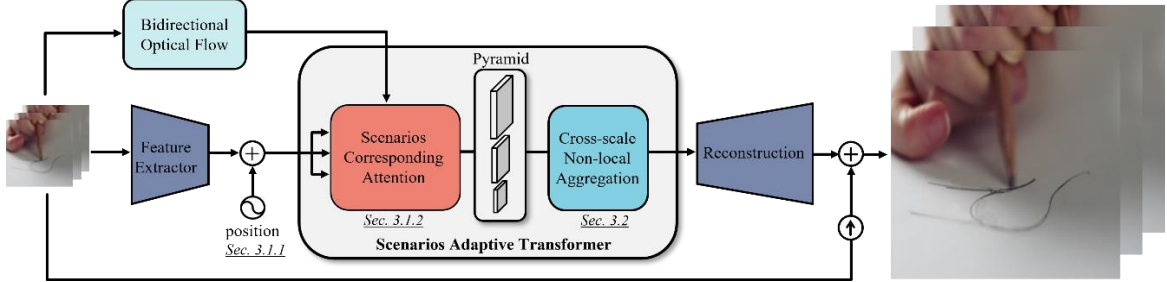

**Figure 1.** The framework of the proposed **Scenarios Adaptive Transformer**. We first capture the features of the LR sequence using a feature extractor. Then, scenarios corresponding attention and cross-scale non-local aggregation model the spatial-temporal representation. Finally, reconstruct the HR video from the representation and upsampling frames.

**3 Our Approach**

In this paper, we propose a novel algorithm called spatial-temporal **S**cenarios **A**daptive **T**ransformer for **V**ideo **S**uper-**R**esolution. The framework of our proposed scenarios adaptation Transformer is illustrated in Figure 1. The SATVSR framework predicts a high-resolution video frame $I_{SR}$, which is close to the ground truth $I_{GT}$. SATVSR consists of a feature extractor, scenarios adaptation Transformer, and a reconstruction module. Specifically, $2N+1$ low-resolution consecutive frames $\mathbf{I}_{LR} = \{I_{LR}^i, i \in [t-N, t+N]\}$ are given, indicating that the central frame $I_{LR}^t$ is a reference frame, and other frames are denoted as neighbor frames. First, we extract features from videos using a series of residual connections. Second, the scenarios adaptation Transformer encoder extracts spatial-temporal information. Third, the cross-scale non-local aggregation module explores similar features from multiple spatial scales. Finally, the reconstruction module recovers high-resolution video frames from the representation.

*3.1 Scenarios Adaptive Transformer*

The structure of the encoder is shown in Figure 2, which has global attention and temporal corresponding attention modules. Given an LR sequence $\mathbf{I}_{LR} = \{I_{LR}^i, i \in [t-N, t+N]\}$, using residual connections to extract features to obtain $F_{[t-N:t+N]}$, the reference frame is denoted as $F_t \in \mathbb{R}^{T \times C \times W \times H}$, we utilize three independent convolutional layers $\phi_q$, $\phi_k$ and $\phi_v$ to capture the spatial features of each video sequence, respectively. Here the size of the convolution kernel is set to 3×3, the stride is set to 1, and the padding is set to 1. We set the stride as s, the sliding local patch size as $P \times P$, and split $F_{[t-N:t+N]}$ into patches $x_{(p,t)} \in \mathbb{R}^{P^2 \times C}, t \in [1,...,T]$ and $p \in [1,...,N]$, where $N = HW/P^2$.

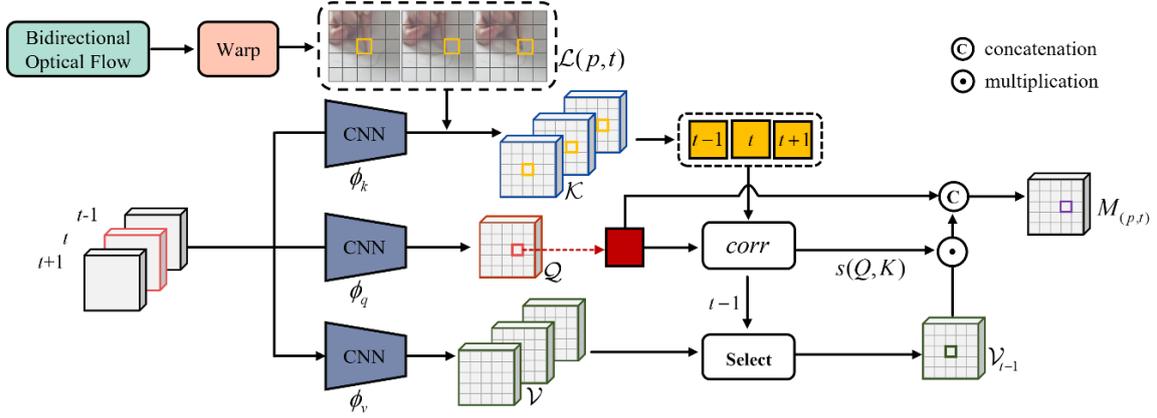

**Figure 2.** Illustration of the scenarios corresponding attention. $\mathcal{Q}, \mathcal{K}, \mathcal{V}$ is generated by three independent convolutional layers, $\mathcal{L}$ represents the label of patches at different periods according to the target frame, *corr* represents the cosine similarity calculation.

*3.1.1 Spatial-temporal positional encoding*
The VSR task requires precise spatial-temporal location information, in order to maintain the location information of each patch, we use 3D fixed location encoding. Specifically, the position encoding includes a temporal positional information, two spatial position information of horizontal and vertical, and its spatial-temporal encoding is expressed as:

$$\text{PE}(p, i) = \begin{cases} \sin(p \cdot \alpha_k), & \text{for } i = 2k, \\ \cos(p \cdot \alpha_k), & \text{for } i = 2k+1, \end{cases} \quad (1)$$

where $\alpha_k = 1/10000^{2k/\frac{d}{3}}$; $p$ is the position in different dimension; $d$ is the number of channels of the feature. We add learnable position encodings to each patch, and get the embedding vector $y_{p,t}^{(0)}$ as the input to the SATransformer:

$$y_{(p,t)} = x_{(p,t)} + \text{PE}.$$

*3.1.2 Scenarios Corresponding attention.*
It has been demonstrated in [5] that utilizing different variants of the Transformer attention mechanism can well model the spatial-temporal dependencies of tokens used in videos. However, in the reconstruction task, the attention mechanism of video is still a challenge. Therefore, we propose an attention module adapted to scenario switching, which adaptively integrates tokens in the same scenario with a small computational cost.

For scenarios corresponding attention, each query/key/value vector is obtained from the feature y with location information through three independent convolutional layers:

$$\mathcal{Q}_{(p,t)}, \mathcal{K}_{(p,t)}, \mathcal{V}_{(p,t)} = \phi_q(y_{(p,t)}), \phi_k(y_{(p,t)}), \phi_v(y_{(p,t)}), \quad (2)$$

where $\phi_{qkv}$ denotes independent convolution computation. It is different from the direct dot product of the $\mathcal{Q}$ vector in the traditional attention mechanism. According to optical flow estimation, we first obtain the spatial location labels $\mathcal{L}(p,t)$ of patches with the same semantic information in the last frame at different times. Second, to reduce the artifacts brought by the introduction of irrelevant information, we compute the cosine similarity to select the most relevant patches along the temporal dimension. When the selected patch comes from the scenario of the non-reference frame, the similarity can reduce the influence of irrelevant information, to achieve the purpose of adaptively utilizing scene information. The formula for this process is:

$$corr(\mathcal{Q}_{(p,t)}, \mathcal{K}_{(p_i,i)}) = \frac{\mathcal{Q}_{(p,t)}}{\|\mathcal{Q}_{(p,t)}\|} \cdot \frac{\mathcal{K}_{(p_i,i)}}{\|\mathcal{K}_{(p_i,i)}\|}, i = 1,...,T, \quad (3)$$

where $t$ represents the time of the restored target frame and $p_i$ represents the location of the patch most relevant to $\mathcal{Q}_{(p,t)}$ selected at $T$ moments using the label $\mathcal{L}(p,t)$. Sorting based on similarity, $s(Q,K)$ is represented as the top 1 value in $corr$. We select the video frame $\mathcal{V}_t$ in $\mathcal{V}$, which is located at the location of the most relevant patch to $\mathcal{Q}_{(p,t)}$. And again through the $\mathcal{L}_{(p,t)}$ label to take out the patch marked in $\mathcal{V}_t$, which is used to supplement the patch information in $\mathcal{Q}$ through the concatenation operation. The process can be expressed as:

$$M_{(p,t)} = \text{SAT}(\mathcal{Q},\mathcal{K},\mathcal{V}) = [\mathcal{Q}_{(p,t)}, s(Q,K) \odot \mathcal{V}_{(p,t)}], \quad (4)$$

where $[\cdot]$ denotes the concatenation operation. the operator $\odot$ denotes multiplication.

*3.2 Cross-scale Non-local Aggregation*
To adapt the model to the multi-scale variation of the content of interest, we propose a cross-scale feature aggregation strategy after SATransforms to extract non-local correspondences from multiple scales. Utilize average pooling to downsample the feature map to get the feature pyramid:

$$M_{(p,t)}^{(l+1)} = AvgPool(M_{(p,t)}^{(l)}), l = \{0,1,2\}. \quad (5)$$

First, select the query patch $m_{(p,t)}^{(0)}$ with position p on the large-scale feature map $M_t^{(0)}$ with level 0. Second, match the patch most relevant to $m_{(p,t)}^{(0)}$ in each level feature. Third, we add self-attention [3] to give different levels of features a weight of whether the feature can be used. Finally, all the features are concatenation and convolved to achieve the purpose of aggregation. The process is formulated as:

$$out = aggr([attn(Mcorr(M_{(p,t)}^{(l)}, m_{(p,t)}^{(0)}))]), \quad (6)$$

where $MCorr(\cdot)$ represents the patch most relevant to $m_{(p,t)}^{(0)}$ in $M_t^{(l)}$, $l = \{1,2,3\}$ represents feature levels of different scales, *attn* is the attention unit, *aggr* is the convolutional layer that aggregates all features together.

**4 Experiments**

*4.1 Datasets and Metrics*
**Scenario Stable Dataset.** To demonstrate the excellent performance of SATVSR, we first compare with other SOTA methods on widely used single scenario datasets (Vid4[11] and Vimeo-90K [6]). For Vid4, it contains four videos of Foliage, Walk, Calendar, and City scenes. For Vimeo-90K, it contains training set and test set (Vimeo-90K-T), where the resolution size of each video is 448×256.
**Scenario Unstable Dataset.** Furthermore, to verify the performance of existing VSR methods on cross scenario videos, we propose a new dataset, Vimeo-90K-fusion (Vimeo-90K-F), based on Vimeo-90K. We set two random numbers, one is to randomly select a sequence in the test set to splice with another sequence, and the other is to randomly select the position of the 7-frame sequence to splice another scene. For a fair comparison with previous work, we use Gaussian filter with a standard deviation of $\sigma = 1.6$ for 4× downsampling to obtain LR images.
**Metrics**. To ensure comparability, we follow the experimental setup [1] and evaluate the quality of images generated by these VSR methods using PSNR and SSIM on the Y channel.

*4.2 Experiment details*
We experiment with two NVIDIA TITAN RTX GPUs, using residual network as feature extraction network. The channel size of the feature maps in the network is set to 64. During training, we decay the learning rate from $2 \times 10^{-4}$ to $10^{-7}$ using the cosine annealing scheme and the Adam optimizer with $\beta_1 = 0.9$ and $\beta_2 = 0.99$. We set the batch size to 2 and the input patch size to 64×64. We use The Charbonnier loss between ground-truth and SR which is defined as: $loss = \sqrt{\|I_{GT} - I_{SR}\|^2 + \varepsilon^2}$. The total number of training iterations is 400K.

**Table 1.** Quantitative comparison (PSNR/SSIM) on Vid4, Vimeo-90-T and Vimeo-90-F dataset for 4× VSR. <span style="color:red">Red</span> and <span style="color:blue">blue</span> indicates the best and the second-best.

|             | Bicubic       | TOFlow[12]    | DUF[5]        | RBPN[8]       | TDAN[4]       | SATVSR(ours)  |
|-------------|---------------|---------------|---------------|---------------|---------------|---------------|
| Vid4        | 21.80/0.5246  | 25.85/0.7659  | 27.38/0.8329  | 27.17/0.8205  | 26.86/0.8140  | 27.81/0.8501  |
| Vimeo-90K-T | 31.30/0.8687  | 34.62/0.9212  | 36.87/0.9447  | 37.20/0.9458  | 36.31/0.9376  | 37.41/0.9489  |
| Vimeo-90K-F | 30.95/0.8626  | 33.12/0.9039  | 23.67/0.7865  | 36.58/0.9412  | -/-           | 36.99/0.9446  |

*4.3 Comparisons with State-of-the-art Methods*

Our proposed SATVSR compares previous SOTAs including, TOFlow [12], DUF [5], RBPN [8], TDAN [4]. The quantitative results in Table 1 are extracted from the original publications. On the Vid4 dataset, SATVSR achieves the highest PSNR and SSIM. In Table 1, it is clearly shown that our method outperforms other methods by at least 0.43dB. On the Vimeo-90K-T dataset, our SATVSR method works better compared to RBPN with nearly 0.21dB enhancement on the Y channel. Several examples are visualized in Figure 3. Previous methods generated blurry images with the inaccurate restoration of texture structure, while our model generated SR results that better preserve image structure and restore details.

In the case of cross scenarios, we find that other methods have heavy performance degradation, among which RBPN, with better performance in a single scenario, also has a performance degradation of 0.62dB. This confirms the disadvantage of exploiting long sequence relationships: insufficient robustness to complex situations. And our design not only has the best effect in a single scenario but also is more robust to cross-scenario data.

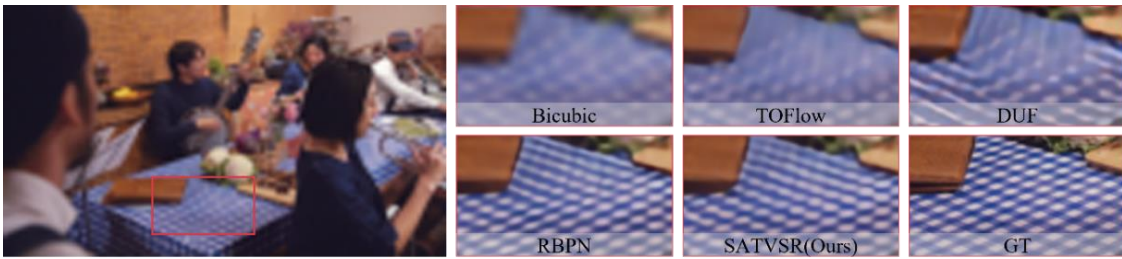

**Figure 3.** Qualitative comparison on Vimeo-90K-T for 4× VSR.

*4.4 Ablation Study*

Table 2 shows the quantitative results of the network utilizing different combinations of modules. Ablation experiments are performed on the proposed SATVSR. First, we do not use optical flow to mark patches, and directly use all patches in space and time as global attention as the "base" model. Secondly, the model that uses the SAT module for the most relevant patch is called the "Base+SAT" model. And the model that finally adds cross-scale feature aggregation is called "Base+SAT+CNA". When the model is not labeled by optical flow, the performance of the model drops significantly, which means that when training with all patches without limitation, it is difficult for the network to learn and utilize scenario information. Secondly, when using a cross-scale non-local aggregation module, the network performance is slightly improved by 0.34dB, and it is confirmed that the module can handle scale changes in video sequences.

**Table 2.** Ablation study results of on the Vid4 [11] dataset.

| Model        | SAT | CNA | PSNR/SSIM     |
|--------------|-----|-----|---------------|
| Base         |     |     | 26.12/0.7813  |
| Base+SAT     | √   |     | 27.47/0.8370  |
| Base+SAT+CNA | √   | √   | 27.81/0.8501  |

**5 Conclusions**

In this work, we introduced a scenario adaptive transformer network for cross-scene video super-resolution tasks, which is another exploration to introduce different attention mechanism transformers into video super-resolution tasks. Specifically, we use pre-trained optical flow to label patches in each video frame and only compute the attention of patches with the same label. According to the attention map, we filter out the most similar patches as the complement of target information. SAT significantly reduces the interference of irrelevant information across scenes on the results and enables the Transformer to extract spatial-temporal information efficiently. Furthermore, the cross-scale feature aggregation module exploits multi-scale information, which further improves the performance of the network. Extensive experiments demonstrate the effectiveness and robustness of our proposed method.